\documentclass[final]{arxiv}

\nolinenumbers
\DOI{xxxxx}
\Year{2026}

\acrodef{ML}{Machine Learning}
\acrodef{MLP}{Multilayer Perceptron}
\acrodef{CNN}{Convolutional Neural Network}
\acrodef{SOTA}{state-of-the-art}
\acrodef{ViT}{Vision Transformer}
\acrodef{HDR}{high dynamic range}
\acrodef{AP}{average precision}
\acrodef{AR}{average recall}
\acrodef{mAP}{mean average precision}
\acrodef{mAR}{mean average recall}
\acrodef{DBH}{diameter at breast height}
\acrodef{UAV}{unmanned aerial vehicle}
\acrodef{SLAM}{simultaneous localization and mapping}
\acrodef{VSLAM}{visual \ac{SLAM}}
\acrodef{HMLS}{handheld mobile laser scanning}
\acrodef{SAM}{Segment Anything Model}
\acrodef{ToF}{Time-of-Flight}
\acrodef{FPS}{frames per second}
\acrodef{SLOAM}{Semantic Lidar Odometry and Mapping}
\acrodef{IoU}{intersection-over-union}
\acrodef{GNSS}{Global Navigation Satellite System}
\acrodef{FLOPs}{floating point operations}
\acrodef{IQR}{interquartile range}

\newcommand{\APfifty}{AP$_{50}$}
\newcommand{\ARfifty}{AR$_{50}$}

\newcommand{\silva}{\mbox{\textit{SilvaScenes}}}
\newcommand{\silvanumimages}{164}
\newcommand{\silvanumspecies}{28}
\newcommand{\silvanumcommonspecies}{20}
\newcommand{\silvanumclasses}{21}
\newcommand{\silvanumtrees}{1421}

\newcommand{\silvanumdomains}{five}

\usepackage{siunitx}
\usepackage{tabularx}
\usepackage{multirow}
\usepackage{subcaption}
\usepackage{booktabs}
\usepackage{comment}

\usepackage[dvipsnames]{xcolor}

\definecolor{navyblue}{RGB}{0,0,128}
\usepackage[
    colorlinks=true,
    linkcolor=navyblue,
    citecolor=navyblue,
    urlcolor=navyblue
]{hyperref}
\usepackage[noabbrev,capitalise]{cleveref}

\usepackage{lipsum}

\newcolumntype{Y}{>{\raggedright\arraybackslash}X} 
\newcolumntype{R}{>{\raggedleft\arraybackslash}X} 
\newcolumntype{Z}{>{\centering\arraybackslash}X} 
\newcolumntype{T}{>{\ttfamily\raggedright\arraybackslash}X} 
\newcolumntype{C}{>{\centering\ttfamily\arraybackslash}X} 

\let\cite\citep

\usepackage{pdflscape}
\usepackage{rotating}

\usepackage[skip=5pt]{parskip}

\begin{document}

\title[\textit{SilvaScenes}]{\textit{SilvaScenes}: Tree Detection and Species Classification \\ from Under-Canopy Images in Natural Forests}

\author[\textit{SilvaScenes}]{David-Alexandre \surname{Duclos}$^{1,\ast}$, William \surname{Guimont-Martin}$^{1}$, Gabriel \surname{Jeanson}$^{1}$,\\ Arthur \surname{Larochelle-Tremblay}$^{1}$, Martine \surname{Lapointe}$^{2}$, Théo \surname{Defosse}$^{3}$, Frédéric \surname{Moore}$^{3}$,\\ Philippe \surname{Nolet}$^{3}$, François \surname{Pomerleau}$^{1}$, and Philippe \surname{Giguère}$^{1}$}

\address{$^{1}$Northern Robotics Laboratory, Université Laval, Québec, QC, G1V 0A6, Canada\\
$^{2}$Département des sciences du bois et de la forêt, Université Laval, Québec, QC, G1V 0A6, Canada\\
$^{3}$Institut des Sciences de la Forêt tempérée, Université du Québec en Outaouais, Ripon, QC, J0V 1V0, Canada\\}

\corres{$^{\ast}$Corresponding author. E-mail: david-alexandre.duclos@norlab.ulaval.ca}

\begin{abstract}
    Interest in forestry automation is growing alongside rapid advances in deep learning.
    In particular, tree detection and taxonomic classification are seen as core tasks required for automating field surveys and forestry equipment.
    These operations must often be performed in under-canopy settings, which pose challenging conditions for perception systems, including heavy occlusion, variable lighting, and dense vegetation.
    Despite this necessity, current work has yet to properly establish the feasibility of simultaneously executing tree detection and taxonomic classification in natural forests, as available datasets primarily focus on urban settings or on a limited number of species.
    To address this gap, we present \silva{}, a benchmark dataset for instance segmentation of tree species from under-canopy images in natural forests.
    Collected across \silvanumdomains{} bioclimatic domains in Quebec, Canada, our dataset features \silvanumtrees{} trees from \silvanumspecies{} species, with segmentation masks for pixel-precise tree trunk detection and fine-grained species annotations from forestry experts.
    We demonstrate the relevance and difficult nature of \silva{} by evaluating modern deep learning approaches, showing that while trunk segmentation is feasible, with a top \ac{mAP} of \SI{69.9}{\%} and \ac{mAR} of \SI{76.4}{\%}, species-aware segmentation remains a significant challenge with an \ac{mAP} and an \ac{mAR} of only \SI{39.2}{\%} and \SI{68.6}{\%}, respectively.
    Alongside additional experiments, we highlight key challenges, namely that species imbalance and tree occlusion figure among the most pressing issues for precise segmentation and identification.
    Meanwhile, higher image resolutions contribute to significant performance gains and will likely prove fundamental to these tasks moving forward.
    Our dataset, source code, and models will be made available at \url{https://github.com/norlab-ulaval/SilvaScenes}.
\end{abstract}

\maketitle

\section{Introduction}

Recent advances in deep learning have paved a promising path for the future of forestry automation \cite{wolk_review_2024}, with anticipated cost reductions, increased worker safety, and more sustainable practices \cite{holzinger2024industry}.
In particular, tree detection and taxonomic classification are seen as key perception tasks for automation and have been well explored through over-canopy solutions \cite{spiers_review_2025}.
Yet, these solutions have been shown to be inaccurate, particularly in densely structured, natural forests \cite{mikita_mapping_2024}.
Some operations, such as plot-level surveys \cite{fassnacht2023remote} and tree harvesting \cite{jelavic_robotic_2022}, can also require these perception tasks to be performed \textit{in situ} and at ground level, highlighting the need for robust, under-canopy solutions.

In natural forests, canopies often form a compact layer, limiting visibility and complicating the association of foliage with corresponding trees.
Tree trunks, and by extension tree bark, are therefore seen as reliable phenotypes for species identification \cite{carpentier_tree_2018}, with the notable advantage of persisting across seasons.
Regarding sensing modalities, recent studies have shown that image-based approaches consistently outperform lidar-based approaches for trunk segmentation at ground level \cite{vidanapathirana_wildscenes_2025}, hinting at the importance of semantic rather than geometric information for forestry tasks.
However, current image-based datasets focus on tree detection \cite{grondin_tree_2023}, taxonomic classification \cite{warner_centralbark_2024}, or a limited combination of both \cite{gade_finnwoodlands_2023}.
Moreover, these datasets are often unrepresentative of natural forests and their inherent complexity.
As such, it remains unclear if current image-based approaches are viable for developing robust perception systems for species-aware tree detection.

\begin{figure}[htbp]
    \centering
    \includegraphics[width=\linewidth]{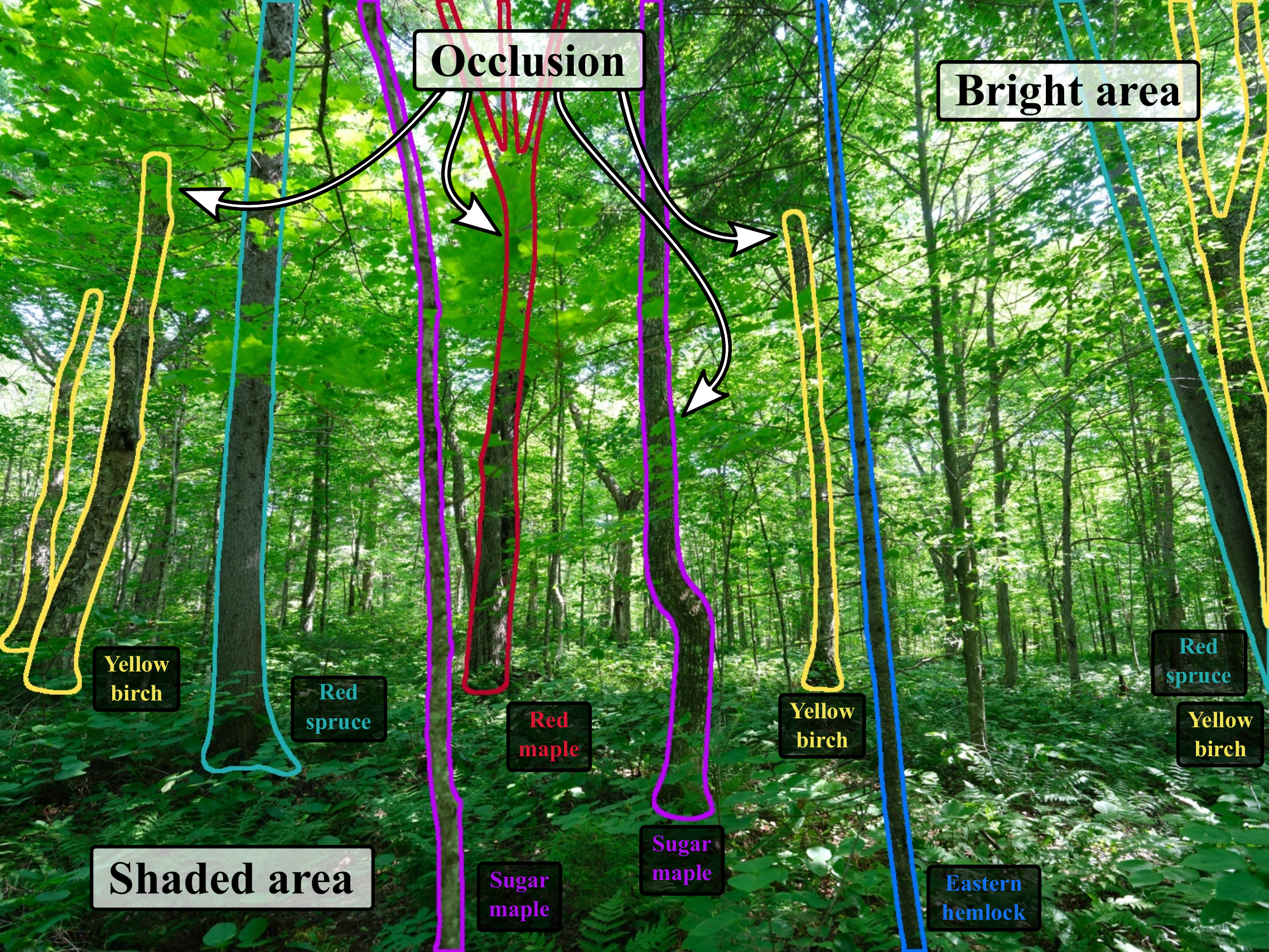}
    \caption{
        Example of an annotated image in our dataset, \mbox{\silva{}}, with high-quality species-wise segmentation masks for tree trunks.
        The image illustrates complex conditions, such as vegetation occlusion and varying lighting, which are frequently found in natural forests.
        Masks are colour-coded by species and drawn with no interior fill to showcase tree bark and occlusion.
    }
    \label{fig:annotation_example}
\end{figure}

To address this gap in the literature, we present \silva{}, a novel benchmark dataset for instance segmentation of tree species from under-canopy images in natural forests.
Our dataset unifies pixel-level trunk detection and fine-grained species classification, with \silvanumtrees{} manually annotated trees from \silvanumspecies{} species.
To capture a diverse and accurately labelled dataset, we collected images across \silvanumdomains{} bioclimatic domains in Quebec, Canada, relying on forestry experts for species identification.
Our dataset features a realistic depiction of natural forests, with highly diverse populations and environments, as well as complex environmental conditions such as heavy occlusion of trees and variable lighting, as can be seen in \cref{fig:annotation_example}.
Furthermore, we demonstrate the utility of our dataset and its challenging nature by benchmarking current deep learning approaches.
Notably, we show that while trunk segmentation in natural forests is feasible, accurate species classification still poses issues.
We publicly release this dataset to encourage the development of automated solutions in forestry and to present a clear measure of the difficulties that deep learning algorithms face in complex, under-canopy settings.
In short, our contributions are:
\begin{itemize}
    \item A benchmark dataset, \silva{}, of under-canopy images for instance segmentation of tree species in natural forests;
    \item An evaluation of current deep learning approaches to demonstrate the challenging nature of our task.
\end{itemize}

\vspace*{-2\baselineskip}

\section{Related work}

In recent years, forestry has increasingly relied on automation for various operations, ranging from inventories and mapping to forest trait estimation.
Over-canopy approaches dominate large-scale surveys, capable of covering broad areas at a relatively low cost.
By contrast, under-canopy approaches capture richer visual information, but must cope with clutter, occlusion, and light variability.
In addition, the choice of sensor modality is of great importance, as it can greatly influence the perception capabilities of autonomous systems.
Finally, some datasets have been proposed to tackle under-canopy tree detection and taxonomic classification.
To contextualize our work within the literature, we review these key points in the following subsections.

\subsection{Over-canopy approaches}

Over-canopy solutions have been extensively studied for regional and national forest inventories by mapping canopy height, segmenting tree crowns, and identifying tree species \cite{spiers_review_2025, zhong_review_2024}.
While approaches have been developed for \ac{DBH} and stem curve estimation, under-canopy approaches have consistently outperformed over-canopy acquisitions relying on \acp{UAV} \cite{hyyppa_comparison_2020}.
Furthermore, over-canopy approaches are misaligned with forestry operations that must be performed at ground level, such as plot-level surveys \cite{fassnacht2023remote}, tree harvesting \cite{jelavic_harveri_2022}, and log grasping \cite{steininger_timbervision_2025}.
Their limitations in canopy penetration, particularly in densely structured forests, often yield lower segmentation and mapping accuracy \cite{mikita_mapping_2024}.
As such, ground-level perception systems remain relevant for automation; the following sections therefore focus on such approaches.

\vspace*{-\baselineskip}

\subsection{Sensor modalities in under-canopy approaches}\label{sec:sensor_modalities}

When it comes to robotics and automation in forestry, two sensing modalities are prevalent: lidar and camera.
Lidar has been widely employed for tree segmentation, geometric trait estimation, and species classification.
For instance, \citet{malladi_tree_2024} used point clouds to estimate the \ac{DBH} and height of trees in forest environments.
Beyond geometry, \citet{wielgosz_segmentanytree_2024} proposed a deep learning method for individual tree segmentation.
Building on this concept, \citet{puliti_benchmarking_2025} benchmarked single-tree species classification, relying on aggregated point clouds constructed from multiple scans.
While their work demonstrates the potential of segmenting fully mapped point clouds, we note that it does not establish the feasibility of online classification during field surveys, which would typically yield single-view and sparser point clouds.
On the other hand, camera-based approaches have been applied to a wide range of tasks.
\citet{lee_testing_2024} demonstrated tree trunk detection from images for real-time under-canopy \ac{UAV} navigation in forests.
Meanwhile, \citet{hristova2025enhancing} proposed a videogrammetry approach for forest mapping and \ac{DBH} estimation of trees, relying on images collected from six fisheye cameras.
Recent studies have also demonstrated image-based solutions for segmenting branches \cite{geckeler_learning_2024} and tree pests \cite{guo2022pest} through dense foliage.
Similarly, approaches have been developed for instance segmentation of logs in harvesting operations, where occlusion introduces important challenges \cite{fortin_instance_2022, steininger_timbervision_2025}.
Incidentally, our \silva{} dataset mirrors these conditions, which are typical of natural forests, providing a difficult benchmark for perception.
Closer to our work, \citet{liu_stereo_2025} proposed an approach for forest mapping and tree genus classification with stereo cameras.
The authors report improved performance when classifying based on aggregated information across multiple images, although their study is limited to five genera in a single urban forest.
Likewise, \citet{liu_classification_2019} developed an approach for semantic segmentation of tree species and stock volume estimation from colour images.
Although promising, the study is limited to identifying four visually distinct species in a single area, thereby limiting its relevance to assessing performance in high-diversity forests.
Importantly, colour cameras are a popular modality, given their high availability and low cost \cite{abreudias2025advances}.
Recent work has shown that approaches for semantic segmentation of tree trunks based on colour images consistently outperform corresponding lidar-based approaches \cite{vidanapathirana_wildscenes_2025, mortimer_goose_2024}, motivating our use of this modality.

\subsection{Image-based datasets for under-canopy approaches}\label{sec:datasets}

A few datasets focus on image-based tree detection in under-canopy environments.
\citet{da_silva_visible_2021} proposed ForTrunkDet, a multi-modal dataset recorded in three Portuguese forests, combining \num{2716} colour and \num{915} thermal images for trunk detection with bounding boxes.
Similarly, \citet{grondin_tree_2023} introduced CanaTree100, a dataset for trunk detection, segmentation, and keypoint estimation.
The dataset contains over \num{920} trees annotated across \num{100} images collected in Quebec, Canada, with instance segmentation masks and keypoints for diameter, felling cut, and inclination.
Compared to bounding boxes, instance segmentation masks provide finer, pixel-level spatial information, which can be of interest when accounting for the natural clutter and highly variable shapes of trees in forests.
Importantly, these datasets do not provide class labels in the ground truth and are thus unsuitable for species classification.

When it comes to taxonomic classification, multiple datasets have been proposed at single-tree level.
\citet{beery_auto_2022} introduced the Auto Arborist dataset for genus classification of \SI{2.6}{M} trees across \num{344} genera, with images sourced from Google Street View.
Although the scale of the dataset is impressive, the images do not pose the same perception challenges as natural forests, as trees are typically isolated and fully visible, providing clear views of the canopy and structure.
Furthermore, distribution of the dataset has officially ended due to maintenance constraints.
On the other hand, \citet{carpentier_tree_2018} proposed BarkNet 1.0, a collection of over \num{23000} close-up images of bark from \num{23} species near Quebec City, Quebec, Canada.
A total of \num{1006} trees are included, along with their \ac{DBH}.
Similarly, \citet{warner_centralbark_2024} developed CentralBark, a dataset with over \num{19000} close-up bark images from \num{4697} trees across \num{25} species native to Indiana, Illinois and Ohio, USA.
In addition to bark images and \ac{DBH}, CentralBark provides bark moisture condition and \ac{GNSS} coordinates.
Both of these works demonstrate the feasibility of accurate species identification from bark images alone, which is relevant for under-canopy perception tasks.
However, approaches that require close-up images of each individual tree sidestep the detection component, thereby reducing their applicability to forestry automation.

\begin{table*}[b]
    \centering
    \caption{
        Comparison of publicly available datasets for tree detection or taxonomic classification from under-canopy images in natural environments.
        \textbf{Images} accounts for colour images, and \textbf{Trees} for unique trees.
    }
    \setlength{\tabcolsep}{6pt}
    \begin{tabularx}{\linewidth}{@{} l Z Z Z R R R @{}}
    \toprule
    \textbf{Dataset} & \multicolumn{3}{c}{\textbf{Task}} & \multicolumn{3}{c}{\textbf{Data}} \\
    \cmidrule(lr){2-4}
    \cmidrule(lr){5-7}
    & \multicolumn{1}{c}{\textbf{Detection}} & \multicolumn{1}{c}{\textbf{Segmentation}} & \multicolumn{1}{c}{\textbf{Classification}} & \multicolumn{1}{c}{\textbf{Images}} & \multicolumn{1}{c}{\textbf{Trees}} & \multicolumn{1}{c}{\textbf{Species}} \\
    \midrule
    ForTrunkDet & \checkmark & -- & -- & 2716 & -- & -- \\
    CanaTree100 & \checkmark & \checkmark & -- & 100 & 920 & -- \\
    FinnWoodlands & \checkmark & \checkmark & \checkmark & 300 & 2562 & -- \\
    BarkNet 1.0 & -- & -- & \checkmark & \num{23000} & 1006 & 23 \\
    CentralBark & -- & -- & \checkmark & \num{19000} & 4697 & 25 \\
    \midrule
    \textbf{\textit{SilvaScenes} (ours)} & \checkmark & \checkmark & \checkmark & \silvanumimages{} & \silvanumtrees{} &  \silvanumspecies{} \\
    \midrule
    \end{tabularx}
    \label{tab:dataset_comparison}
\end{table*}

Research that simultaneously addresses both tree detection and taxonomic classification in under-canopy settings is limited.
\citet{yang_urban_2023} created the Tree Dataset of Urban Street (TDoUS), which includes classification and segmentation of trees and their components, such as trunks, crowns, and fruits.
A total of 29 species are presented in the trunk images taken across ten cities in China.
However, visibility on urban streets is high, obstruction is minimal, and resource competition among trees is nonexistent.
This dataset therefore poorly translates to natural forests, which develop with minimal human intervention \cite{forrester_spatial_2014}.
In natural forests, \citet{gade_finnwoodlands_2023} created FinnWoodlands, a dataset for semantic, instance, and panoptic segmentation from snowy trails in Finland, with a total of \num{2562} annotated trees across 300 images.
Importantly, the authors classify three tree genera, but do not distinguish between species.
In addition, snowy environments have high visual contrast and low vegetation occlusion, which can ease segmentation and classification.
As a result, current datasets are insufficient for benchmarking image-based instance segmentation of tree species in high-diversity, natural forests.
We present a comparison with existing under-canopy image datasets in \cref{tab:dataset_comparison}.
While previous datasets lack species labels or are limited to classifying single trees, \silva{} is the first dataset to offer densely annotated images and high-quality instance segmentation masks for precise detection and fine-grained species classification of trees in natural forests.

\vspace*{-0.5\baselineskip}

\section{The \silva{} dataset}

To advance the deployment of autonomous systems in forestry settings, we introduce the \silva{} dataset, taken across \mbox{Quebec}, Canada, in June and July 2025.
Our dataset is representative of natural forests, with a high diversity of environments and tree species, as well as adverse conditions such as heavy clutter, visual occlusion, and variable lighting.
\Cref{tab:tree_species} shows the distribution of tree species in \silva{}, presented taxonomically.
Notably, the dataset contains a total of \silvanumspecies{} tree species, with \silvanumtrees{} unique trees.
Many species are present across multiple bioclimatic domains, increasing both the environmental and intraspecific diversity of our dataset.
Given the high species imbalance, which is typical of natural forests \cite{nasiri_using_2025}, experiments were mainly conducted on our \silvanumcommonspecies{} most common species by setting a minimum threshold of 16 specimens per species.
To ensure rigorous data collection and annotation and to properly direct future efforts, we propose a set of guidelines tailored to the complexity of forests.
We see our dataset as a robust benchmark and, importantly, a building block toward the development of perception systems for natural environments.
The following sections describe the equipment, bioclimatic domains, and guidelines used to create \silva{}.

\onecolumn
\begin{landscape}
    \begin{table}[htbp]
    \centering
    \caption{
        Tree species present in the dataset \silva.
        We describe their taxonomy, followed by the number collected in each bioclimatic domain.
        Rows are sorted alphabetically by taxonomy.
        Common names are sourced from Canada’s National Forest Inventory's Tree Species List \cite{canadian_forest_service_canadas_2014}.
    }
    \setlength{\tabcolsep}{3pt}
    \renewcommand{\arraystretch}{1.1645}
    \begin{tabularx}{\linewidth}{@{} *{3}{l} *{2}{Y} *{6}{c} @{}}
    \toprule
    & \textbf{Family} & \textbf{Genus} & \textbf{Species (Latin)} & \textbf{Species (Common)} & \multicolumn{6}{c}{\textbf{Number of trees per bioclimatic domain}} \\
    &  &  &  &  & \textbf{SM-BH} & \textbf{SM-YB} & \textbf{SM-BW} & \textbf{BF-YB} & \textbf{BF-WB} & \textbf{Total} \\
    \midrule
    \multirow{24}{*}{\rotatebox{90}{Deciduous}} & \multirow{3.5}{*}{\textit{Betulaceae}} & \multirow{2}{*}{\textit{Betula}} & \textit{alleghaniensis} Britt. & Yellow birch & 6 & 4 & 24 & 37 & -- & 71 \\
     &  &  & \textit{papyrifera} Marsh. & White birch & 20 & 21 & 2 & 9 & 30 & 82 \\
    \cmidrule(lr){3-11}
     &  & \multirow{1}{*}{\textit{Ostrya}} & \textit{virginiana} (Mill.) K. Koch & Ironwood & 29 & 8 & -- & -- & -- & 37 \\
    \cmidrule(lr){2-11}
     & \multirow{3.5}{*}{\textit{Fagaceae}} & \multirow{1}{*}{\textit{Fagus}} & \textit{grandifolia} Ehrn. & American beech & 71 & 3 & 43 & 3 & -- & 120 \\
    \cmidrule(lr){3-11}
     &  & \multirow{2}{*}{\textit{Quercus}} & \textit{bicolor} Willd. & Swamp white oak & 3 & -- & -- & -- & -- & 3 \\
     &  &  & \textit{rubra} L. & Red oak & 21 & 25 & -- & -- & -- & 46 \\
    \cmidrule(lr){2-11}
     & \multirow{2.5}{*}{\textit{Juglandaceae}} & \multirow{1}{*}{\textit{Carya}} & \textit{cordiformis} (Wangenh.) K. Koch & Bitternut hickory & 37 & -- & -- & -- & -- & 37 \\
    \cmidrule(lr){3-11}
     &  & \multirow{1}{*}{\textit{Juglans}} & \textit{cinerea} L. & Butternut & 1 & -- & -- & -- & -- & 1 \\
    \cmidrule(lr){2-11}
     & \multirow{1}{*}{\textit{Malvaceae}} & \multirow{1}{*}{\textit{Tilia}} & \textit{americana} L. & Basswood & 41 & 4 & -- & -- & -- & 45 \\
    \cmidrule(lr){2-11}
     & \multirow{3}{*}{\textit{Oleaceae}} & \multirow{3}{*}{\textit{Fraxinus}} & \textit{americana} L. & White ash & 32 & 4 & -- & -- & -- & 36 \\
     &  &  & \textit{nigra} Marsh. & Black ash & 1 & -- & -- & 2 & -- & 3 \\
     &  &  & \textit{pennsylvanica} Marsh. & Red ash & 3 & -- & -- & -- & -- & 3 \\
    \cmidrule(lr){2-11}
     & \multirow{2}{*}{\textit{Rosaceae}} & \multirow{2}{*}{\textit{Prunus}} & \textit{serotina} Ehrh. & Black cherry & 18 & -- & -- & -- & -- & 18 \\
     &  &  & \textit{pensylvanica} L. & Pin cherry & 1 & -- & -- & -- & -- & 1 \\
    \cmidrule(lr){2-11}
     & \multirow{2}{*}{\textit{Salicaceae}} & \multirow{2}{*}{\textit{Populus}} & \textit{grandidentata} Michx. & Largetooth aspen & 21 & -- & -- & -- & -- & 21 \\
     &  &  & \textit{tremuloides} Michx. & Trembling aspen & -- & 2 & -- & 9 & 16 & 27 \\
    \cmidrule(lr){2-11}
     & \multirow{3}{*}{\textit{Sapindaceae}} & \multirow{3}{*}{\textit{Acer}} & \textit{pensylvanicum} L. & Striped maple & 13 & -- & 8 & -- & -- & 21 \\
     &  &  & \textit{rubrum} L. & Red maple & 45 & 47 & 51 & 24 & -- & 167 \\
     &  &  & \textit{saccharum} Marsh. & Sugar maple & 162 & 3 & 51 & 27 & -- & 243 \\
    \cmidrule(lr){2-11}
     & \multirow{1}{*}{\textit{Ulmaceae}} & \multirow{1}{*}{\textit{Ulmus}} & \textit{americana} L. & White elm & 1 & -- & -- & -- & -- & 1 \\
    \midrule
    \midrule
    \multirow{10}{*}{\rotatebox{90}{Coniferous}} & \multirow{1}{*}{\textit{Cupressaceae}} & \multirow{1}{*}{\textit{Thuja}} & occidentalis L. & Eastern white-cedar & -- & 60 & -- & -- & -- & 60 \\
    \cmidrule(lr){2-11}
     & \multirow{8.45}{*}{\textit{Pinaceae}} & \multirow{1}{*}{\textit{Abies}} & \textit{balsamea} (L.) Mill. & Balsam fir & -- & 30 & 8 & 46 & 166 & 250 \\
    \cmidrule(lr){3-11}
     &  & \multirow{1}{*}{\textit{Larix}} & \textit{laricina} (Du Roi) K. Koch & Tamarack & -- & 2 & -- & -- & -- & 2 \\
    \cmidrule(lr){3-11}
     &  & \multirow{3}{*}{\textit{Picea}} & \textit{glauca} (Moench) Voss & White spruce & -- & -- & -- & 3 & 17 & 20 \\
     &  &  & \textit{mariana} (Mill.) B.S.P. & Black spruce & -- & -- & -- & -- & 17 & 17 \\
     &  &  & \textit{rubens} Sarg. & Red spruce & -- & -- & -- & 17 & -- & 17 \\
    \cmidrule(lr){3-11}
     &  & \multirow{1}{*}{\textit{Pinus}} & \textit{strobus} L. & Eastern white pine & 3 & -- & -- & -- & -- & 3 \\
    \cmidrule(lr){3-11}
     &  & \multirow{1}{*}{\textit{Tsuga}} & canadensis (L.) Carrière & Eastern hemlock & 33 & 22 & -- & 1 & -- & 56 \\
    \midrule
    \midrule
     &  &  &  & Unknown & 4 & 2 & 5 & 1 & 1 & 13 \\
    \bottomrule
    \multicolumn{11}{l}{\footnotesize{
        \emph{Legend:} 
        \textbf{SM}: sugar~maple;
        \textbf{BF}: balsam~fir;
        \textbf{BH}: bitternut~hickory;
        \textbf{BW}: basswood;
        \textbf{YB}: yellow~birch;
        \textbf{WB}: white~birch.
    }}\\
    \end{tabularx}
    \label{tab:tree_species}
    \end{table}
\end{landscape}
\twocolumn

\subsection{Equipment}

Camera use in under-canopy environments presents unique challenges, such as high dynamic range and depth of field trade-offs \cite{gamache_exposing_2024}. 
We chose to conduct our off-trail data collections in a handheld manner, following typical acquisition protocols \cite{gade_finnwoodlands_2023, vidanapathirana_wildscenes_2025}.
This approach allows for better control of motion blur, camera angle, and camera parameters, while avoiding challenges with robot navigation, to focus on the core perception challenges of our task.
We used a Fujifilm~GFX100S camera, featuring a \num{43.8}~$\times$~\SI{32.9}{mm} sensor with a resolution of \SI{102}{MP} (i.e.,~\num{11648}~$\times$~\SI{8736}{px}).
The lens was a Fujifilm~GF23mmF4~R~LM~WR, with a \SI{99.9}{\degree} diagonal field of view, offering a balance between wide-angle coverage and minimal radial distortion.
Furthermore, our large lens enables better light capturing and a greater depth of field.
In practice, we set our aperture size to around f/\num{6.4} and our shutter speed to approximately \num{1}/\SI{50}{s}, and minimize the ISO number.
The result is an extended depth of field with minimal blur and noise and adequate gain.
To account for the prohibitive scaling of current deep learning solutions with respect to image resolution \cite{liu_swin_2021, tan_efficientnetv2_2021}, we downsample our images to \SI{1.6}{MP} (i.e.,~\num{1456}~$\times$~\SI{1092}{px}), which is akin to previous works \cite{gade_finnwoodlands_2023, grondin_tree_2023}.

\subsection{Bioclimatic domains}

\begin{figure*}[htbp]
    \centering
    \includegraphics[width=\linewidth]{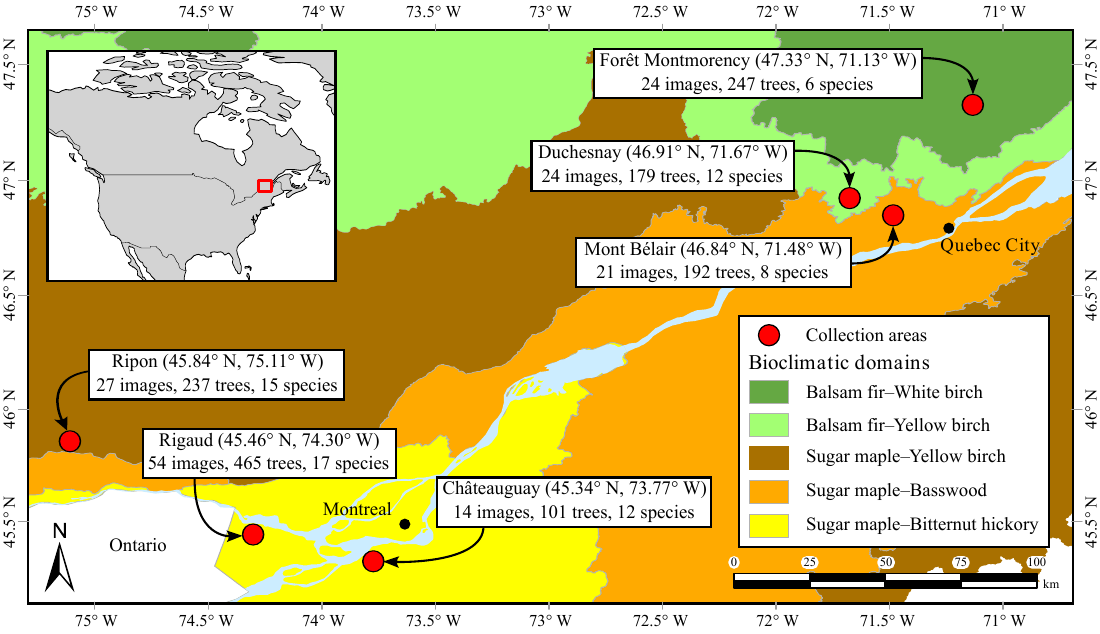}
    \caption{
        Map of the data collection areas for \silva{}.
        Images are taken across multiple sites in these areas, most of which span several kilometres.
    }
    \label{fig:dataset_map}
\end{figure*}

We present a map of the data collection areas for \silva{} in \cref{fig:dataset_map}.
Our images are distributed across \silvanumdomains{} bioclimatic domains, which are ecoregions defined by Quebec's Ministry of Natural Resources and Forests as end-of-succession territories with unique vegetation and climate \cite{ministere_des_ressources_naturelles_et_des_forets_zones_2022}.
The \emph{Sugar~maple--Bitternut~hickory} is a small domain in the south temperate zone, characterized by highly fertile soils and deciduous species. 
A significant amount of our data was collected in this domain, as it has the highest tree species diversity in Quebec, with 48 different species.
The \emph{Sugar~maple--Basswood} surrounds the previous domain, with a cooler climate and a higher presence of coniferous species.
The \emph{Sugar~maple--Yellow~birch} extends from the Canadian Shield of Témiscamingue to the St.~Lawrence Valley, and is characterized by the decline of many species commonly found in the previous domain.
In addition, clear-cuts in this domain often lead to stands dominated by red maple and white birch.
The \emph{Balsam~fir--Yellow~birch} is a transitional domain in the north temperate zone, characterized by low-altitude plains and a reduced presence of deciduous species.
Finally, the \emph{Balsam~fir--White~birch} is a southern boreal domain with both plains and mountainous terrains, composed almost exclusively of coniferous species.

\begin{figure*}[htbp]
    \centering
    \includegraphics[width=0.9\linewidth]{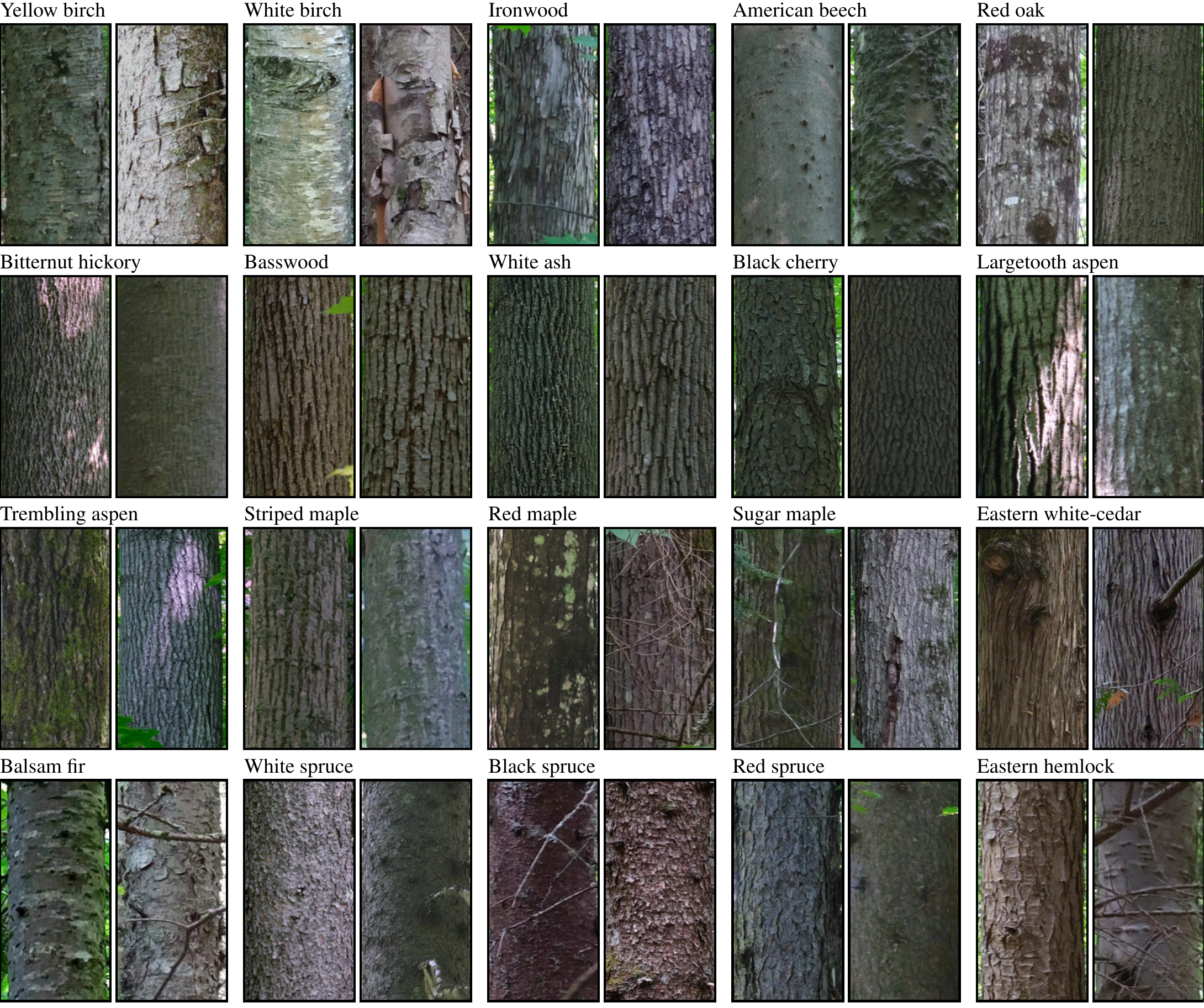}
    \caption{
        Examples of tree barks from the most common species in \silva{}, demonstrating the high level of inter- and intra-species diversity and similarity.
        Images were extracted from our full \SI{102}{MP} images.
        Species are ordered alphabetically by taxonomy.
        Brightness and green hue were adjusted to account for variations in scene illumination and occasional colour bleeding from dense canopies.
    }
    \label{fig:crops_mosaic}
\end{figure*}

By collecting data in these \silvanumdomains{} bioclimatic domains, our dataset includes a rich diversity of both tree species and forest settings.
Furthermore, our data were collected at multiple sites for a mixture of inter- and intra-domain diversity.
This collection strategy is essential, as the appearance of species can greatly vary across different environments.
We demonstrate the high bark-level diversity and similarity across our most common species in \cref{fig:crops_mosaic}, highlighting one of the many difficulties of identifying trees from under-canopy images.
For example, we can observe that maples share visual characteristics with many other species, namely red oak, ironwood, and both aspens.
On the other hand, multiple species have high intraspecific diversity based on various factors such as age and size, complicating their identification.
Finally, our data were mainly collected off-trail to fully represent the clutter and complexity of natural forests and to better align with forestry operations.
Properly representing the diversity and complexity of forests is crucial to developing robust solutions that can generalize across different environments.

\subsection{Data collection}\label{sec:data_collection}

In addition to collecting data in different bioclimatic domains, we sought to capture a broad diversity of scenes, representative of the many conditions that may be encountered in natural forests.
As such, we established the following collection guidelines:
\begin{enumerate}
    \item Images are taken with an emphasis on species and environmental diversity.
    We vary the number of trees per image, their position with respect to the camera, and prioritize less common species.
    \item Images are mainly collected off-trail to fully represent the complexity of natural forests, such as species competition, heavy occlusion, and low lighting \cite{forrester_spatial_2014}.
    \item We avoid capturing an individual tree, labelled or not, more than once across all images.
    Furthermore, images are taken across multiple sites in a given collection area.
    Enforcing these criteria is crucial, as duplicated trees or repeating elements can bias experiments through data leakage \cite{robert_tree_2020, dale2002spatial}.
\end{enumerate}

\begin{figure*}[htbp]
    \centering
    \begin{subfigure}[t]{0.25\linewidth}
        \centering
        \includegraphics[height=120pt]{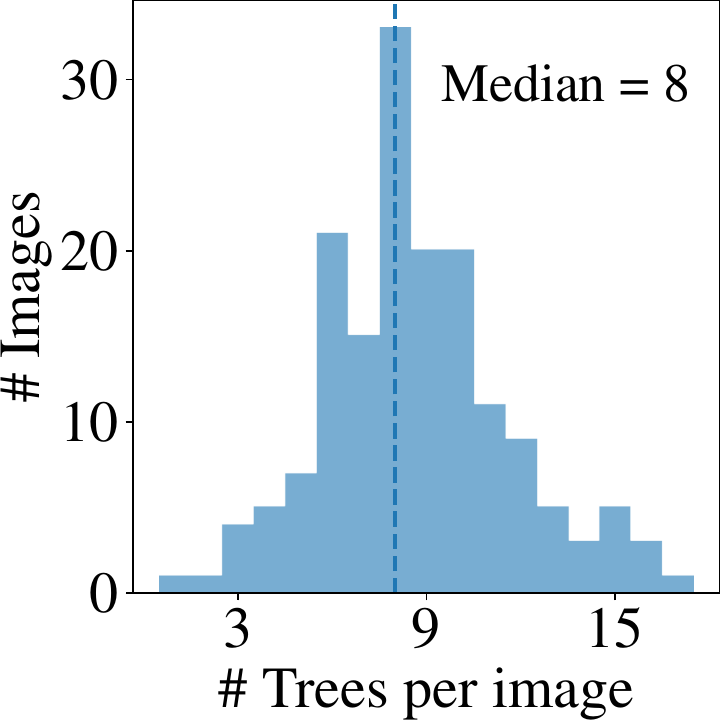}
        \caption{}
        \label{fig:nb_trees_distribution}
    \end{subfigure}\hfill
    \begin{subfigure}[t]{0.25\linewidth}
        \centering
        \includegraphics[height=120pt]{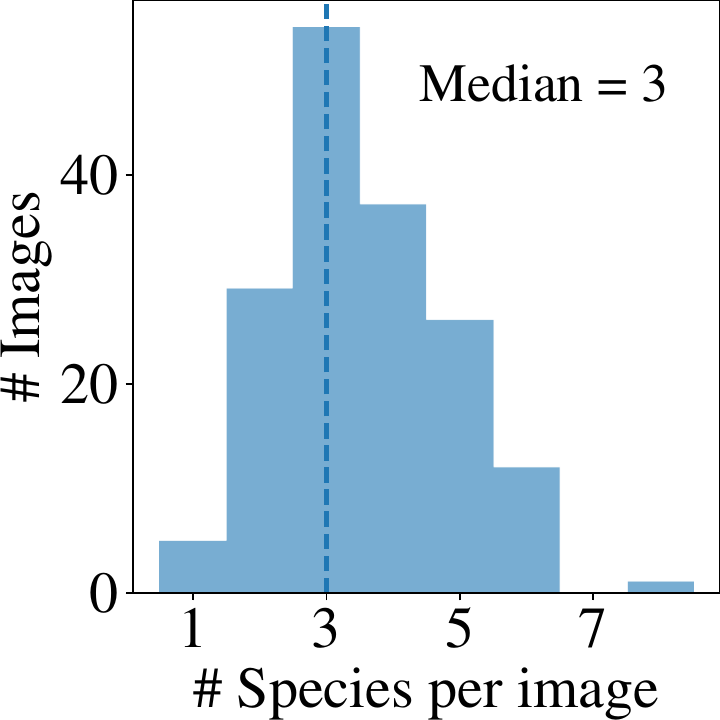}
        \caption{}
        \label{fig:nb_species_image}
    \end{subfigure}\hfill
    \begin{subfigure}[t]{0.25\linewidth}
        \centering
        \includegraphics[height=120pt]{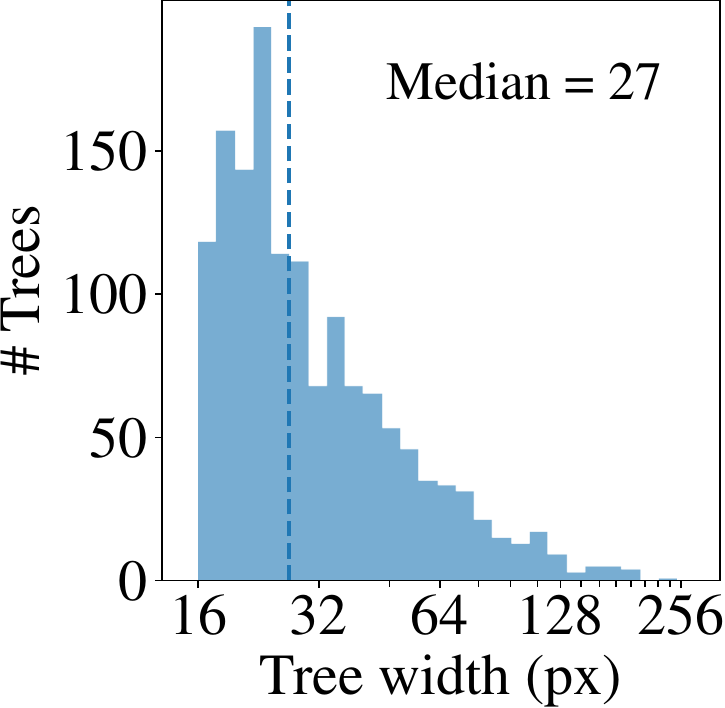}
        \caption{}
        \label{fig:tree_width_distribution}
    \end{subfigure}\hfill
    \begin{subfigure}[t]{0.25\linewidth}
        \centering
        \includegraphics[height=120pt]{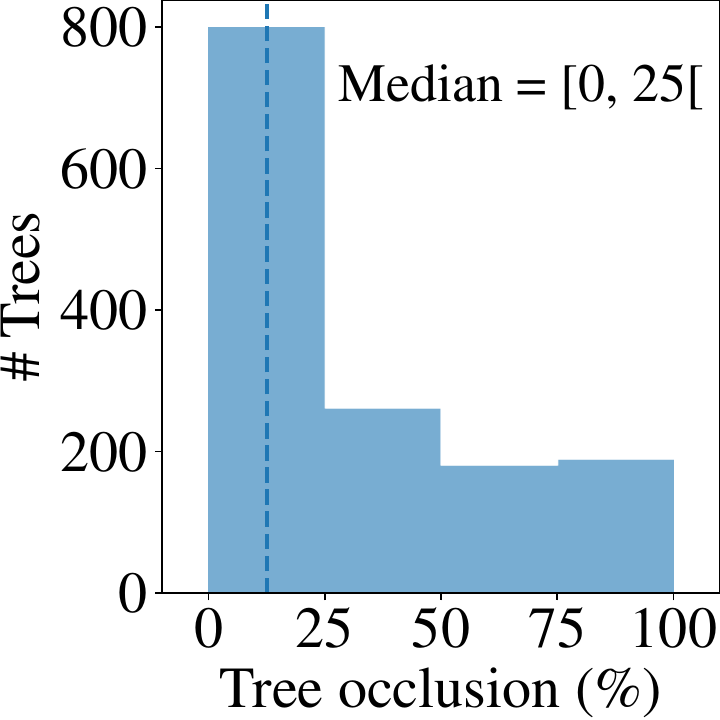}
        \caption{}
        \label{fig:tree_occlusion_distribution}
    \end{subfigure}
	\caption{
        Statistics of our \silva{} dataset.
        (a) Number of trees per image.
        (b) Number of species per image.
        (c) Log-scale distribution of tree width in our downsampled images.
        (d) Approximate distribution of tree occlusion across our images.
    }
	\label{fig:dataset_statistics}
\end{figure*}

To demonstrate the diversity of tree species and environments, \cref{fig:nb_trees_distribution} shows the distribution of the number of trees per image, while \cref{fig:nb_species_image} illustrates the number of distinct species per image.
Both distributions follow Gaussian trends, with median values of eight trees and three species per image, respectively.
In addition, \cref{fig:tree_width_distribution} displays the distribution of tree widths, measured as the median width across a tree's height.
Finally, as noted by \citet{grondin_tree_2023}, measuring tree occlusion in images is difficult.
While these authors elected to estimate occlusion and its impact on tree detection through synthetically generated images, we chose instead to visually approximate occlusion in our dataset using four levels ranging from \num{0} to \SI{100}{\%}, in increments of \SI{25}{\%}.
The resulting distribution of tree occlusion can be seen in \cref{fig:tree_occlusion_distribution}.
In total, nearly half of our trees are occluded by at least \SI{25}{\%}, while almost one out of seven trees is occluded by \SI{75}{\%} or more.
This highlights the prevalence of occlusion in under-canopy images and the challenging nature of our task, and will be of interest in our later studies.

\subsection{Data annotation}\label{sec:data_annotation}

Our images were annotated with class labels and instance segmentation masks for individual trees.
There are many challenges when annotating trees in forest environments, such as trees coming in various shapes and sizes, and heavy obstruction from vegetation.
To properly direct our efforts, we established the annotation guidelines below.

\begin{enumerate}
    \item Human identification of tree species from images is difficult \cite{fiel_automated_2011, carpentier_tree_2018}.
    As such, ground truth for most of the data was obtained \textit{in situ} by forestry experts, who could rely on bark, leaves, shoots, cones, shapes, and environmental factors to identify each tree.
    \item Trees are annotated with instance segmentation masks for finer spatial localization.
    Masks are limited to trunks, as branches and foliage are difficult to annotate and are not necessary for forestry operations such as harvesting \cite{grondin_tree_2023}.
    \item Obstructed segments of trunks are labelled if their shape can be inferred from the image.
    Trunk sections are labelled if obstructed by branches or foliage, but not if overlapped by another segmented trunk.
    This labelling practice is akin to previous work on segmenting occluded branches \cite{geckeler_learning_2024}.
    \item If a trunk forks below breast height (\SI{1.3}{m}), each section is considered a separate tree, following the specifications from the \citet{canadian_forest_service_canadas_2008}.
    \item To ensure sufficient visibility for both segmentation and classification, small trees are not annotated.
    A tree is considered small if the median width across its height is less than \SI{16}{px} in our downsampled images.
    \Cref{fig:tree_width_distribution} demonstrates the resulting distribution of tree widths, which closely follows a log-normal distribution.
    Our chosen threshold provides a balance between annotation completeness and tree visibility.
    \item Trees that cannot be reliably identified due to heavy damage, disease, or death are grouped under the Unknown class.
\end{enumerate}

Interestingly, recent works have looked into using the Segment Anything family of models (SAM) for automatic annotation of images \cite{kirillov_segment_2023, wolk_enhancing_2025, grondin2024prompt}.
However, we found that the instance segmentation masks produced by these models were often noisy, imprecise, and misaligned with our annotation guidelines, in most part due to occlusion from surrounding vegetation.
As such, our masks were drawn and revised by human annotators, leveraging the full-resolution \SI{102}{MP} images for increased precision.
Nevertheless, given a larger amount of data, these tools could provide an interesting trade-off between quality and quantity of annotations \cite{khoreva2017weakly}.

\section{Benchmark experiments}

Following the prevalent use of deep learning in forestry automation \cite{ouaknine_openforest_2025}, we conduct benchmark experiments with widely used instance segmentation models.
In the next section, we detail the neural network architectures, training setup, and performance metrics used in our experiments.

\subsection{Network architectures}

Deep learning approaches for image-based tasks are typically based on either \acp{CNN} or \acp{ViT}.
\acp{CNN} rely on convolutional filters that scan across an image to detect patterns such as colours, shapes, and textures.
These filters are learned during training and are computationally efficient, making \acp{CNN} practical for many applications.
However, because these filters operate over relatively small regions, \acp{CNN} are inherently restricted in their ability to model long-range relationships between distant parts of an image \cite{dosovitskiy_image_2021}.
On the other hand, \acp{ViT} rely on attention mechanisms that can attend to all regions of an image simultaneously, allowing them to capture intricate, global relationships, which can be advantageous in complex environments.
However, this flexibility comes at a cost, as \acp{ViT} typically require significantly more training data and computation \cite{liu_swin_2021}.
Given these trade-offs, we opted to benchmark both approaches.

For \acp{CNN}, we chose YOLO-based architectures, which have been applied for detection and segmentation tasks in forestry \cite{wolk_enhancing_2025, gyawali_tree_2025}.
Specifically, we used YOLOv11\footnote{\url{https://github.com/ultralytics/ultralytics}} and YOLOv12 \cite{tian_yolov12_2025}, with the latter adopting a hybrid approach with attention operations.
For \acp{ViT}, we chose Mask2Former \cite{cheng_masked-attention_2022} with a Swin Transformer \cite{liu_swin_2021} backbone, a combination which performs well on different forestry-related tasks \cite{vidanapathirana_wildscenes_2025, fortin_instance_2022}.
Swin Transformer relies on a more efficient hierarchy-based attention mechanism, scaling linearly instead of quadratically.
Furthermore, we experimented on small and large variants of each model to benchmark potential trade-offs between computational efficiency and performance.
The Small YOLOv11 and YOLOv12 models are around \SI{9}{M} parameters, while the \mbox{X-Large} variants are around \SI{60}{M}, approximately 6$\times$ larger.
As for Mask2Former, the Swin-Small and Swin-Large variants are of \SI{69}{M} and \SI{216}{M} parameters, respectively, a size factor of around 3$\times$.
We note that Mask2Former with Swin-Small and the X-Large YOLO models have a similar parameter size, which will be of interest when comparing architectures.
Lastly, we further assessed the best-performing model to characterize its performance.

\subsection{Training details}

For YOLO, we used the implementations from Ultralytics.\footnotemark[2]
Changes were made to add support for non-contiguous instance segmentation masks, such as trees with multiple sections or gaps.
For Mask2Former, we used the implementation from the Transformers library from HuggingFace \cite{wolf_transformers_2020}.
All models are implemented in PyTorch and are pre-trained for instance segmentation on the general-purpose COCO dataset \cite{fleet_microsoft_2014}.
Each model was trained with its native data augmentation pipeline.
To mitigate the impact of class imbalance, we replaced Mask2Former's cross-entropy loss for classification with focal loss \cite{tsung-yi_lin_focal_2017}, which is also used in YOLO.
Hyperparameters were tuned for each experiment through Bayesian hyperparameter search with Weights \& Biases.\footnote{\url{https://wandb.ai}}

Due to the prohibitive scaling of prevalent deep learning solutions \cite{liu_swin_2021, tan_efficientnetv2_2021}, images and masks were downsampled to \SI{1.6}{MP}, a resolution which is akin to previous works \cite{gade_finnwoodlands_2023, grondin_tree_2023}.
Given the limited size of our dataset, we followed a stratified five-fold cross-validation approach for each of our experiments.
Images were automatically split into five folds, while ensuring that each fold had approximately \SI{20}{\%} of each species' trees.
For proper training and evaluation, we set a minimum requirement of 16 specimens per species.
Although this threshold is arbitrary, we note that the resulting distribution of species is comparable to BarkNet 1.0 \cite{carpentier_tree_2018}.
A total of eight species did not meet this threshold and were combined with Unknown trees into a class named Other, similarly to \citet{gade_finnwoodlands_2023}.
Thus, we conduct our experiments on \silvanumclasses{} classes.

\subsection{Performance metrics}

For instance segmentation, we measure performance with \ac{AP} and \ac{AR}, which reflect the quality of predicted masks and their associated classes, using the standard COCO evaluation metrics.
The \ac{AP} metric is mainly relevant when measuring false positives (i.e., commission errors), while \ac{AR} is used for false negatives (i.e., omission errors).
The \APfifty{} and \ARfifty{} metrics enforce a minimum \ac{IoU} of \SI{50}{\%} between predictions and ground truth.
This can be of interest when accounting for occlusion-based errors or ambiguity, and may also offer sufficient precision for certain forestry tasks.
The \ac{AR} metrics attempt to match up to 100 predictions with the ground truth, which can be useful to measure a model's ability to detect all trees or species within a scene.
Meanwhile, the \ac{mAP} and \ac{mAR} metrics are the mean of \acp{AP} and \acp{AR} over \ac{IoU} thresholds ranging from \SI{50}{\%} to \SI{95}{\%}, sampled at \SI{5}{\%} intervals, providing a more comprehensive assessment of mask quality.
All metrics are reported as an average across classes (i.e., macro-average) to account for class imbalance.
We consider the number of parameters and \ac{FLOPs} of each model, as these metrics are of interest in low-compute mobile systems, and \ac{FPS} for real-time applications.

\section{Results and discussions}\label{sec:results}

In this section, we first evaluate our models for instance segmentation of tree species on our \silva{} dataset.
To systematically isolate the factors influencing model performance, we include an evaluation of species-agnostic instance segmentation of trees.
We then leverage the best-performing model for further experiments.
We present a qualitative study of model predictions alongside an analysis of model confusion for species classification.
Finally, we evaluate the impact of tree occlusion and image resolution on our tasks.

\subsection{Main results}

Results for instance segmentation of tree species and trees are presented in \cref{tab:results}, with additional computational metrics provided in \cref{tab:computational_metrics}.
Starting with tree segmentation, Mask2Former with Swin-Large consistently achieves the highest mean across all metrics, obtaining an \APfifty{} of \SI{90.8}{\%} and an \ARfifty{} of \SI{98.1}{\%}.
Given a minimum segmentation \ac{IoU} of \SI{50}{\%}, this means that more than \SI{90}{\%} of the tree detections are valid, while more than \SI{98}{\%} of trees are correctly detected.
Interestingly, most models achieve comparable performance for \APfifty{} and \ARfifty{}, highlighting that tree detection is a simple task, even for very small models.
For \ac{mAP} and \ac{mAR}, the Mask2Former models obtain comparable performance, while the YOLO models achieve consistently lower performance.
Mask2Former with Swin-Large achieves an \ac{mAP} of \SI{69.9}{\%} and an \ac{mAR} of \SI{76.4}{\%}, showcasing the difficulty of high-quality segmentation in natural forests.
For operations such as harvesting and robot navigation, where precision may be key for an accurate grasp or a correct traversability assessment, this performance may be insufficient.
In comparison, \citet{grondin_tree_2023} achieve \SI{60.0}{\%} for \ac{mAP}, \SI{87.2}{\%} for \APfifty{}, \SI{65.2}{\%} for \ac{mAR}, and \SI{91.5}{\%} for \ARfifty{} with their CanaTree100 dataset.
A few factors could explain our improved performance, such as differences in neural network architectures, higher image quality, and annotation methodology.

\begin{table*}[htbp]
    \centering
    \caption{
        Results for instance segmentation on \silva{}.
        Metrics are reported as macro-average percentages ± standard deviation across classes using a five-fold cross-validation strategy.
        Best results are in \textbf{bold}, and second-best results are \underline{underlined}.
    }
    {
    \setlength{\tabcolsep}{5pt}
    \begin{tabularx}{\linewidth}{Y Y *{8}{c}}
        \toprule
        \textbf{Architecture} & \textbf{Backbone} & \multicolumn{4}{c}{\textbf{Species segmentation (\%)}} & \multicolumn{4}{c}{\textbf{Tree segmentation (\%)}} \\
        \cmidrule(lr){3-6}
        \cmidrule(lr){7-10}
        & & \textbf{mAP} & \textbf{\APfifty{}} & \textbf{mAR} & \textbf{\ARfifty{}} & \textbf{mAP} & \textbf{\APfifty{}} & \textbf{mAR} & \textbf{\ARfifty{}} \\
        \midrule
        Mask2Former & Swin-Small & 29.5$_{\pm 5.2}$ & 39.2$_{\pm 6.8}$ & 59.5$_{\pm 3.9}$ & 76.2$_{\pm 3.8}$ & \underline{67.6}$_{\pm 2.1}$ & \underline{89.7}$_{\pm 1.4}$ & \underline{74.2}$_{\pm 1.8}$ & \underline{97.7}$_{\pm 0.8}$ \\
        & Swin-Large & \textbf{39.2}$_{\pm 2.4}$ & \textbf{51.2}$_{\pm 2.8}$ & \textbf{68.6}$_{\pm 1.9}$ & \textbf{89.0}$_{\pm 1.6}$ & \textbf{69.9}$_{\pm 1.4}$ & \textbf{90.8}$_{\pm 1.8}$ & \textbf{76.4}$_{\pm 0.8}$ & \textbf{98.1}$_{\pm 0.9}$ \\
        YOLOv11 & Small & 28.3$_{\pm 3.6}$ & 40.5$_{\pm 5.3}$ & 52.2$_{\pm 3.2}$ & 72.9$_{\pm 4.1}$ & 59.7$_{\pm 1.8}$ & 87.2$_{\pm 2.2}$ & 69.8$_{\pm 1.4}$ & 96.5$_{\pm 0.7}$ \\
        & X-Large & \underline{35.2}$_{\pm 4.3}$ & \underline{48.0}$_{\pm 5.4}$ & \underline{64.1}$_{\pm 1.7}$ & \underline{85.9}$_{\pm 2.2}$ & 60.6$_{\pm 1.5}$ & 87.9$_{\pm 1.4}$ & 69.7$_{\pm 1.2}$ & 96.3$_{\pm 1.1}$ \\
        YOLOv12 & Small & 24.2$_{\pm 3.1}$ & 35.4$_{\pm 5.1}$ & 53.3$_{\pm 0.8}$ & 75.7$_{\pm 1.7}$ & 60.4$_{\pm 3.5}$ & 88.0$_{\pm 1.8}$ & 69.8$_{\pm 2.3}$ & 96.6$_{\pm 1.3}$ \\
        & X-Large & 31.4$_{\pm 1.7}$ & 42.7$_{\pm 2.4}$ & 62.7$_{\pm 2.5}$ & 83.8$_{\pm 2.6}$ & 58.6$_{\pm 3.0}$ & 85.6$_{\pm 1.4}$ & 69.0$_{\pm 2.2}$ & 95.1$_{\pm 0.9}$ \\
        \bottomrule
    \end{tabularx}
    }
    \label{tab:results}
\end{table*}

\begin{table}[htbp]
    \centering
    \caption{
        Computational metrics for the model architectures and backbones used in our experiments.
        \Ac{FPS} is reported on an NVIDIA RTX 4090 GPU with BF16-mixed precision, and includes pre- and post-processing time.
        Best results are in \textbf{bold}, and second-best results are \underline{underlined}.
    }
    {
    \setlength{\tabcolsep}{3pt}
    \begin{tabularx}{\linewidth}{l l r r R}
        \toprule
        \textbf{Architecture} & \textbf{Backbone} & \textbf{Params\hspace{0.5mm}(M)} & \textbf{FLOPs\hspace{0.5mm}(B)} & \textbf{FPS} \\
        \midrule
        Mask2Former & Swin-Small & \num{68.8} & \num{313.0} & \num{7.0} \\
        & Swin-Large & \num{216.0} & \num{868.0} & \num{4.7} \\
        YOLOv11 & Small & \underline{9.4} & \textbf{35.5} & \textbf{57.7} \\
        & X-Large & \num{56.9} & \num{319.0} & \num{33.0} \\
        YOLOv12 & Small & \textbf{9.3} & \underline{35.7} & \underline{51.8} \\
        & X-Large & \num{59.1} & \num{325.0} & \num{20.6} \\
        \bottomrule
    \end{tabularx}
    }
    \label{tab:computational_metrics}
\end{table}

Moving now to tree species segmentation, Mask2Former with Swin-Large again achieves the highest mean across all metrics, with an \ac{mAP} of \SI{39.2}{\%}, an \APfifty{} of \SI{51.2}{\%}, an \ac{mAR} of \SI{68.6}{\%}, and an \ARfifty{} of \SI{89.0}{\%}.
Importantly, these results highlight a clear trend: performance is significantly worse with the inclusion of a species classification component.
The highest degradation is observed among the \ac{mAP} and \APfifty{} metrics, which suffer a \num{30.7} and \num{39.6} point loss, respectively.
Additionally, there is a significant discrepancy between metrics, as \ac{AP} is considerably worse than \ac{AR}.
This gap is primarily driven by the high visual similarity between certain species, which leads to model confusion.
When faced with species ambiguity, the models seem to attempt multiple predictions rather than refrain from predicting, leading to high \ac{AR} but low \ac{AP}.
It is noteworthy that YOLOv11 with an X-Large backbone achieves comparable performance for these species-wise metrics, but that this comes at the cost of a higher standard deviation across folds.
In contrast to the previous task, Mask2Former with Swin-Small does not achieve comparable performance for tree species segmentation.
Interestingly, YOLOv11 X-Large surpasses YOLOv12 X-Large in most metrics, hinting that attention mechanisms may not be beneficial for our tasks.

Crucially, across each architecture, there is a clear demonstration of model size playing an important role.
However, when accounting for standard deviation, no architecture is consistently best across our results.
For instance, Mask2Former with Swin-Small has a comparable parameter size to the YOLOv11 X-Large model, but each is stronger at a different task.
If hardware is not constrained and a low \ac{FPS} is acceptable, Mask2Former with Swin-Large is the clear winner.
Otherwise, both Mask2Former with Swin-Small and YOLOv11 X-Large offer an interesting trade-off between performance and speed, which may be advantageous in applications where real-time performance is critical.
Importantly, these results demonstrate that while tree detection is a simple task, precise segmentation and fine-grained species classification from under-canopy images in natural forests are very challenging.
Indeed, our highest results, particularly across the \ac{mAP} and \ac{mAR} metrics, leave ample room for future improvements.

\subsection{Qualitative results}

\begin{figure*}[htbp]
    \centering
    \includegraphics[width=\linewidth]{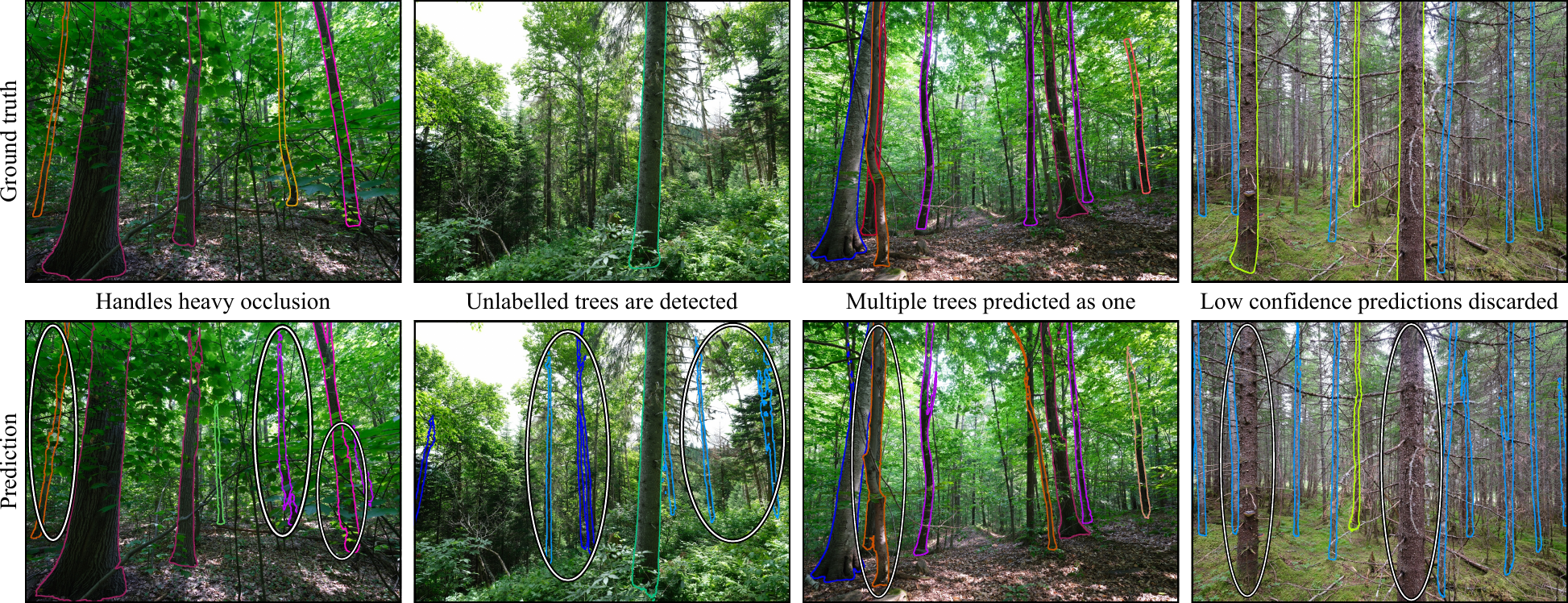}
    \caption{
        Examples of instance segmentation predictions using Mask2Former with Swin-Large.
        Mask contours are colour-coded by species, and points of interest are highlighted with ellipses.
    } 
    \label{fig:prediction_examples}
\end{figure*}

Across most metrics, the Mask2Former model with Swin-Large achieves the strongest performance; we thus select it for further analysis.
Qualitative results of instance segmentation of tree species are presented in \cref{fig:prediction_examples}.
As highlighted in the first example, the model demonstrates the ability to handle heavy occlusion.
Even when leaves or branches hide significant portions of a trunk, the model can accurately delineate its position, which is vital for robust perception systems in natural forests.
However, it is clear that in extreme cases, occlusion may impact the model's ability to accurately segment or altogether detect a tree, or identify its species.
We note that image quality is impacted by colour bleeding under dense canopies, which alters the white balance towards green hues, as observed by \citet{carpentier_tree_2018}.
This shift in white balance may affect the model's ability to detect or identify trees based on their characteristic colours.
In the second example, the model detects multiple trees that are not included in our ground truth, which is consistent with the findings of \citet{grondin_tree_2023}.
This behaviour may be exacerbated by the use of scale-based data augmentations, used to promote scale invariance, which could hinder the model's ability to replicate our annotation methodology.
It should be noted that these predictions are not inherently bad but are misaligned with our ground truth.
We showcase in the third example that, in rare instances, multiple trees can be predicted as one.
Relying on distance or size measurements through the use of depth images or depth estimation models could mitigate these issues \cite{geckeler_learning_2024, wolk_forestry_2025}, at the expense of increased computation and complexity.
Finally, in the fourth example, some predictions are discarded due to low confidence, which is a measure of the model's uncertainty when classifying trees.
Introducing a separate detection confidence, independent of classification uncertainty, could solve this and be of particular interest for mapping forests.

\subsection{Confusion matrix}

\begin{figure*}[htbp]
    \centering
    \includegraphics[width=0.75\linewidth]{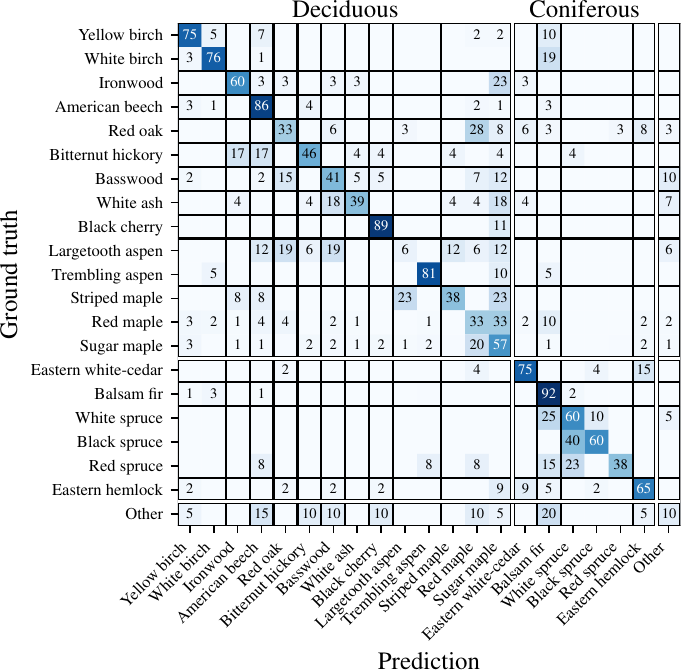}
    \caption{
        Confusion matrix for tree species classification using Mask2Former with Swin-Large over five folds.
        Predictions and ground truth are matched using a geometry-based matcher.
        Results are row-normalized and expressed in percentages.
        Species are split into deciduous, coniferous, and Other, and grouped to highlight intra-genus confusion.
    }
    \label{fig:confusion_matrix}
\end{figure*}

Next, we examine classification performance with the confusion matrix shown in \cref{fig:confusion_matrix}.
For this study, we match instance segmentation predictions with ground truth using a geometry-based matcher, which computes optimal bipartite matches based on \ac{IoU}.
We impose a minimum \ac{IoU} of \SI{50}{\%}, along with a prediction confidence threshold of 0.5, which corresponds to the default setting for Mask2Former.
Interestingly, confusion between deciduous and coniferous species is relatively low, with an accuracy of \SI{92.5}{\%}.
Many deciduous species were occasionally misidentified as red maples or sugar maples, which can have smooth, rugged, or cracked bark depending on various factors such as age and environment.
The frequent prediction of red maples and sugar maples is further worsened by their prevalence, with a similar issue occurring with our most abundant species, balsam fir.
These misclassifications can be attributed to the species imbalance shown in \cref{tab:tree_species}, which is typical of natural forests \cite{nasiri_using_2025}.
Although some measures were taken to alleviate this imbalance, it remains an important issue.
Further use of training techniques, such as class-balanced sampling or more advanced data augmentations, could improve results \cite{puliti_benchmarking_2025}.
At one of our collection sites in the \emph{Balsam~fir--White~birch} bioclimatic domain, some balsam firs exhibited bark detachment, a condition likely associated with resource competition between trees.
This bark loss, which is similar to that observed on yellow and white birches, may have contributed to the confusion between these species.
Although our dataset contains only 18 specimens of black cherry, precision on this species is surprisingly strong.
Conversely, the largetooth aspen has the lowest precision with 21 specimens.
The trembling aspen was one of the only deciduous species present in the \textit{Balsam Fir} bioclimatic domains, likely explaining why they were comparatively easy to identify.
This highlights the importance of properly presenting the full range of forest diversity, which we aim to improve upon in future work.
It is notable that maple and spruce have high intra-genus confusion rates.
While it is difficult to distinguish these species from bark alone, reliable classification has been achieved on the BarkNet 1.0 dataset \cite{carpentier_tree_2018}.
In comparison, our images are subject to harsher environmental conditions and have significantly lower bark-level resolution.
Additionally, we note that BarkNet 1.0 was mainly collected around a single city, which may have contributed to the data being less representative of the full range of species' visual diversity.
Finally, the Other class, which groups Unknown trees and less common species, is very challenging and is akin to open-set or background recognition issues \cite{nasiri_using_2025}.
A recent study by \citet{low2025openset} suggests that it may be possible to reliably identify such species, although this identification would require specialized training objectives.

\subsection{Impact of tree occlusion}

\begin{figure}[htbp]
    \centering
    \includegraphics[width=\linewidth]{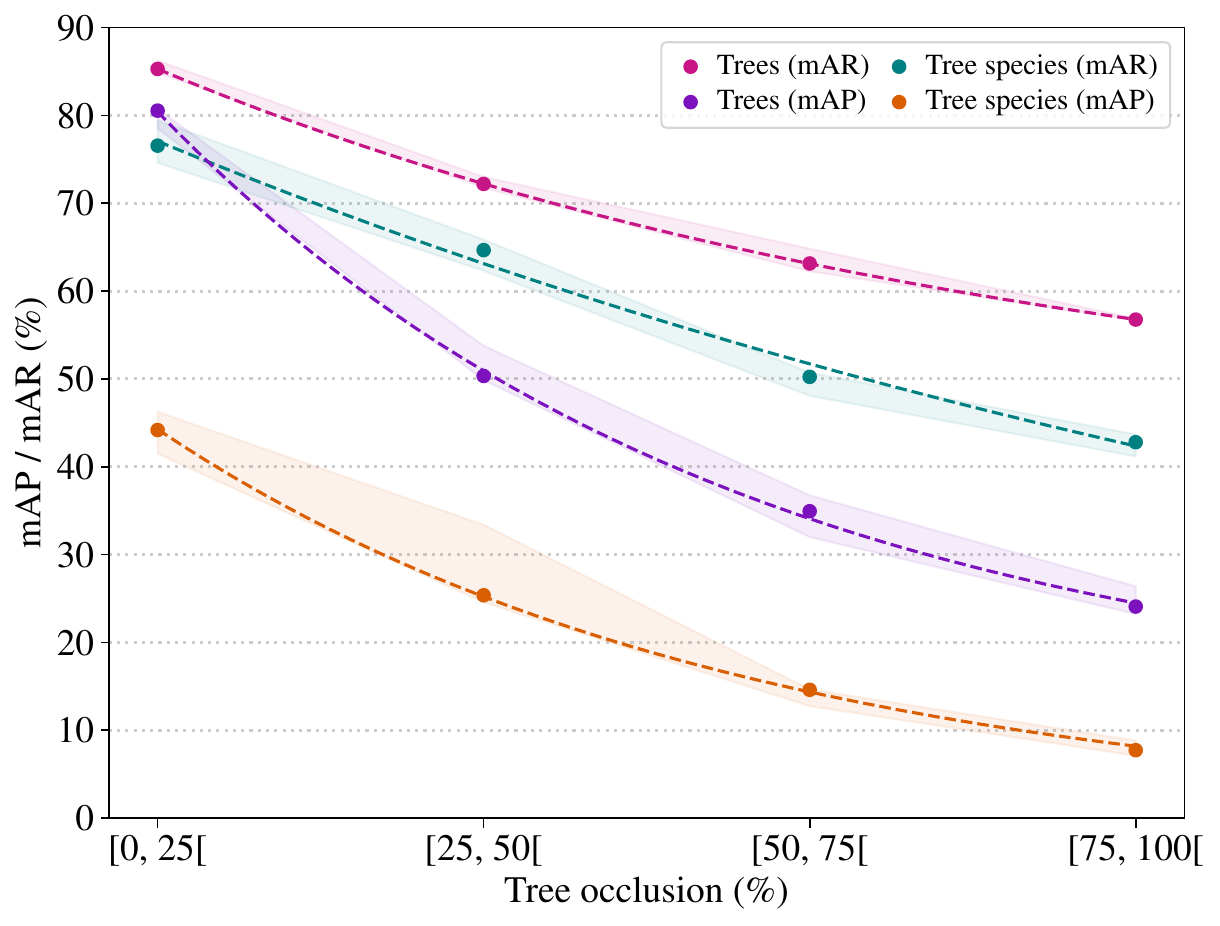}
    \caption{
        Impact of tree occlusion on tree and species segmentation using Mask2Former with Swin-Large.
        Bands show the \ac{IQR} over five folds.
    }
    \label{fig:performance_occlusion}
\end{figure}

Next, we present a study on the impact of tree occlusion on instance segmentation of trees and species.
To compute results for each occlusion level, we follow the COCO protocol for attribute-based evaluation.
As illustrated in \cref{fig:performance_occlusion}, higher levels of occlusion result in important performance degradation across all metrics.
For \ac{mAR}, performance goes down by around \num{12} points for both tasks, even just moving from $[0, 25[$ to $[25, 50[$.
Meanwhile, \ac{mAR} drops by an astonishing \num{28.5} and \num{33.8} points for trees and species, respectively, when shifting from $[0, 25[$ to $[75, 100[$.
As such, occlusion makes trees harder to segment accurately or even to detect entirely.
As a reminder, tree bark is among the most important features for species classification in under-canopy images; occlusion from leaves and branches can make bark texture harder to recognize, thus increasing the difficulty of this task.
Even under very heavy occlusion, the \ac{mAR} for tree segmentation stays relatively high.
On the other hand, the impact of occlusion seems greater for \ac{mAP}, as occlusion leads to significant ambiguity for tree shape.
Moving from $[0, 25[$ to $[25, 50[$ occlusion leads to an important drop of \num{30.2} points, while going to $[75, 100[$ drops by a total of \num{56.5} points.
Overall, as expected, occlusion is an important issue in under-canopy images.
Importantly, we argue that this trend could be alleviated through certain improvements, most notably by increasing image resolution.

\subsection{Impact of image resolution}

\begin{figure}[htbp]
    \centering
    \includegraphics[width=\linewidth]{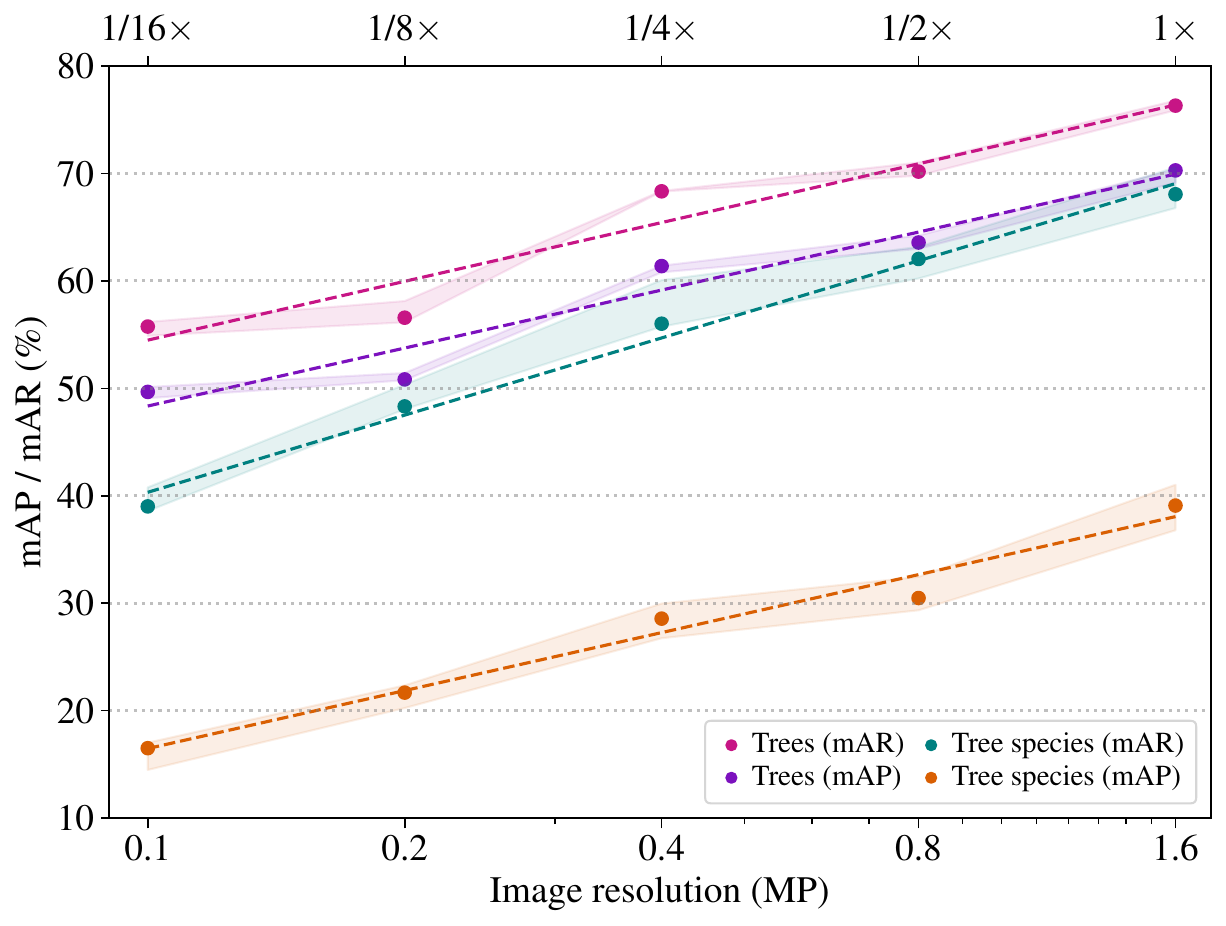}
    \caption{
        Impact of image resolution on tree and species segmentation using Mask2Former with Swin-Large.
        Bands show the \ac{IQR} over five folds.
        Note that image resolution is in log scale.
    }
    \label{fig:performance_resolution}
\end{figure}

In light of these results, we present an additional study on the impact of image resolution on instance segmentation of trees and species.
As a reminder, although cameras such as ours are capable of reaching resolutions of over \SI{100}{MP}, prevalent deep learning solutions typically operate at lower resolutions of around \SI{1.6}{MP}, as higher-resolution images introduce important computational challenges for both model training and deployment.
To study the importance of this design choice, we downsample our images by steps of factor two, from our baseline of \SI{1.6}{MP} down to \SI{0.1}{MP}.
For each image resolution, we train and evaluate a separate model following the same five-fold cross-validation strategy.
As shown in \cref{fig:performance_resolution}, image resolution displays an important trend toward improved performance across both tasks.
For tree segmentation, both \ac{mAP} and \ac{mAR} increase by approximately 20 points between \SI{0.1}{MP} and \SI{1.6}{MP}, which is indicative of lower ambiguity for tree shape.
Meanwhile, \ac{mAP} and \ac{mAR} for tree species closely follow a power law, increasing respectively by \num{7.2} and \num{5.4} points each time image resolution is doubled.
Most importantly, there is no clear indication of a plateau across any metric.
In a best-case scenario, assuming these trends do not yet plateau, using our full \SI{102}{MP} images (i.e., 64$\times$ our baseline resolution) could yield an estimated \ac{mAP} of \SI{70.4}{\%} for tree species segmentation, while \ac{mAR} could reach near-perfect performance.
As established in our previous study, occlusion is an important challenge for both segmentation and identification; it is therefore promising that image resolution seems to counteract this detrimental factor.
These results underline the fundamental role that image resolution could play in our tasks, and possibly across many more moving forward.

\section{Conclusion and future work}

In this paper, we presented \silva{}, a benchmark dataset for instance segmentation of tree species from under-canopy images in natural forests.
Collected across Quebec, Canada, \silva{} captures \silvanumspecies{} species in \silvanumimages{}, featuring high-quality annotations for \silvanumtrees{} unique trees.
While existing datasets typically focus on either detection or classification, or on simple conditions, our dataset unifies these tasks within realistic, highly challenging conditions which are typical in natural forests, such as heavy clutter, severe visual occlusion, and high species diversity.
By proposing this dataset and establishing rigorous protocols for our data collection and labelling, we aim to address the critical lack of data in forestry contexts and provide a foundational framework to guide the creation of future datasets.

Using the \silva{} dataset, we benchmarked modern deep learning solutions for instance segmentation, evaluating the performance of Mask2Former, YOLOv11, and YOLOv12 models.
In our experiments, the best model, Mask2Former with a Swin-Large backbone, achieves impressive performance for tree trunk segmentation, with an \APfifty{} of \SI{90.8}{\%} and a \ARfifty{} of \SI{98.1}{\%}, highlighting that even in highly complex forest scenes, tree detection remains feasible.
By contrast, the model obtains an \ac{mAP} of \SI{69.9}{\%} and an \ac{mAR} of \SI{76.4}{\%}, demonstrating the difficulty of precise tree segmentation under complex environmental conditions.
Crucially, the inclusion of species classification further exposes the challenging nature of our dataset, as results for \ac{mAP} and \APfifty{} plummet to \SI{39.2}{\%} and \SI{51.2}{\%}, respectively.

Further studies highlight several critical issues.
Although models can locate trunks through heavy occlusion, segmentation quality and species identification are significantly impacted.
For instance, \ac{mAP} for tree segmentation drops from \SI{80.6}{\%} for minimally occluded trees to \SI{24.1}{\%} for highly occluded trees, while species-level \ac{mAP} falls from \SI{44.2}{\%} to just \SI{7.7}{\%}.
Furthermore, the absence of distance and size measurements in colour images occasionally leads to the false detection of small or distant trees, a behaviour misaligned with our annotation guidelines.
Our evaluation of model confusion for species classification showcases that deciduous trees are rarely confused with coniferous trees.
However, model confusion for species is considerably worse, highlighting that this will be a key issue moving forward.
Among the challenges are maples and spruces, which share visual characteristics with other species and among themselves, and an imbalanced representation of species, which is inherent to natural forests.
Crucially, our results indicate that the performance of prevalent deep learning models is constrained by the standard practice of using lower-resolution images, as both tree and species segmentation scale consistently with resolution, with no clear indication of a plateau at our baseline of \SI{1.6}{MP}.

To overcome these limitations, a promising direction for future work is to leverage very-high-resolution images of \SI{100}{MP} or higher, which are being increasingly studied in remote sensing and biomedical sciences \cite{bakhtiarnia_efficient_2024}.
Our images have already been collected at \SI{102}{MP}, which will greatly facilitate this transition.
Importantly, the use of these very-high-resolution images is not trivial, as it introduces significant computational challenges and trade-offs to consider for both training and deploying perception systems.
Meanwhile, we plan to expand \silva{}, with a focus on new and underrepresented species, in addition to more environmental diversity.
Moreover, we wish to include information about a tree's approximate age, size, and health, which could help overcome previously highlighted misidentification issues and be of high interest for forestry operations.
As demonstrated in prior work, the use of hierarchical classification techniques based on taxonomy \cite{mu_globalgeotree_2026} and of aggregated information across multiple images \cite{carpentier_tree_2018, liu_stereo_2025} could also be of interest to address species confusion.
Another approach we intend to explore is the use of multi-modal or semi-supervised approaches, which have shown great promise for alleviating the costs of data collection and annotation \cite{ouaknine_openforest_2025}.
Finally, we plan to deploy our perception pipeline onto mobile robots to assess its robustness and viability for real-time automation in complex, natural forests.

\section*{Data availability}
Our \silva{} dataset, source code, and models will be available on our GitHub repository upon publication of this paper.

\section*{Funding}
This work was supported by the Fonds de recherche du Québec Doctoral Research Scholarship [2006691]; the Department of National Defence/Natural Sciences and Engineering Research Council of Canada Discovery Grant Supplements [DGDND-04741-2022]; and the Canada Foundation for Innovation Fund [39709].
We gratefully acknowledge the support of the NVIDIA Corporation with the donation of a Quadro RTX 8000 GPU, which was used for some of our experiments.

\bibliographystyle{arxiv}
\bibliography{references}

\end{document}